\documentclass[journal]{IEEEtran}

\usepackage{hyperref}
\usepackage{cite}
\usepackage{authblk}
\usepackage{graphicx}
\usepackage{amsmath}
\usepackage{multicol,lipsum}
\usepackage{cleveref}
\usepackage{xcolor}
\ifCLASSINFOpdf

\begin{document}
\UseRawInputEncoding

\title{Classification and Segmentation of Pulmonary Lesions in CT Images Using a Combined VGG-XGBoost Method, and an Integrated Fuzzy Clustering-Level Set Technique}

\author[1] {Niloofar Akhavan Javan\thanks{\textbf{Niloofar Akhavan Javan} is a Master graduate with from the Department of Computer Engineering, Khayyam University, Iran. Email: \href{niloofar.akhavan.javan@gmail.com}{niloofar.akhavan.javan@gmail.com}}}

\author[2] {Ali Jebreili\thanks{\textbf{Ali Jebreili} is an Assistant Professor of Computer Engineering at Khayyam University, Iran. Email: \href{alii.jebreili@gmail.com}{alii.jebreili@gmail.com}}}

\author[3] {Babak Mozafari\thanks{\textbf{Babak Mozafari} is a Master graduate from the Department of Computer Engineering, Khayyam University, Iran. Email: \href{ba.mozafari@gmail.com}{ba.mozafari@gmail.com}}}

\author[4] {Morteza Hosseinioun\thanks{\textbf{Morteza Hosseinioun} is a recently graduate with a Master's from the Department of Computer Engineering, School of Science and Engineering, Sharif University of Technology, Iran. \textcolor{blue}{He is also a corresponding author for this work.} Email: \href{hosseinioun@ce.sharif.edu}{hosseinioun@ce.sharif.edu}}}

\author[5] {S. AmirAli Gh. Ghahramani\thanks{\textbf{S. AmirAli Gh. Ghahramani} is an Assistant Professor of Computer Engineering department at Sharif University of Technology, Kish International Campus, Iran. Email: \href{ghahramani@ce.sharif.edu}{ghahramani@ce.sharif.edu}}}

\affil[1, 2, 3]{Department of Computer Engineering, Khayyam University}
\affil[4, 5]{School of Engineering and Science, Department of Computer Engineering, Sharif University of Technology}

\markboth{Library of Cornell University, arXiv, April~2022}
{Shell \MakeLowercase{\textit{et al.}}: Bare Demo of IEEEtran.cls for IEEE Journals}
\maketitle

\begin{abstract}
Given that lung cancer is one of the deadliest illnesses, early identification and diagnosis are critical to preserving a patient's life. However, lung illness diagnosis is time-intensive and requires the expertise of a pulmonary disease specialist, subject to a significant rate of inaccuracy. Our objective is to design a system capable of accurately detecting and classifying lung lesions and segmenting them in CT-scan images. The suggested technique extracts features automatically from the CT-scan image and then classifies them using Ensemble Gradient Boosting methods. Finally, if a lesion is detected in the CT-scan image, it is segmented using a hybrid approach based on Fuzzy Clustering and Level Set \cite{LI20111}. To train and test our models we gathered a dataset that included CT images of patients residing in Mashhad, Iran. Finally, the results indicate 96\% accuracy within this dataset. This approach may assist clinicians in diagnosing lung abnormalities and avoiding potential errors.
\end{abstract}
\begin{IEEEkeywords}
Pulmonary Lesion Classification and Segmentation, Deep Learning, VGG Convolutional Neural Networks, XGBoost, Level Set Methods.
\end{IEEEkeywords}

\IEEEpeerreviewmaketitle

\section{Introduction}

Cancer is a group of diseases that are characterized by the growth of uncontrollable abnormal cells. If the spread of the abnormal cell is not controlled, it can lead to death. However, the disease's cause is unknown for many cancers, especially those that occur during childhood. Many factors that cause cancer are known, including lifestyle factors such as smoking, overweight and unmodifiable factors, such as mutations hereditary, hormonal, and immune conditions. These risk factors may be associated simultaneously or continuously to initiate and/or promote cancer growth. In 2020, the American Cancer Society’s lung cancer measures in the United States indicated roughly 228,820 new cases of lung cancer amongst 116,300 in men and 112,520 in women and approximately 135,720 mortality from lung cancer in between 72,500 in men and 63,220 in women \cite{americansoci2}. Lung cancer is the second most common cancer in women and men (irrespective of skin cancer) and is one of the most important causes of cancer deaths among men and women. The mortality rate from lung cancer is higher than colon, breast, and prostate cancer altogether. However, the detection of small lung nodules from volumetric CT-scans is also tricky, and for this reason, many CAD tools are designed to compensate for this problem \cite{americansoci3, survival2020}. If lung cancer is detected at an early stage, when it is small and has not spread yet, a person has a greater chance of living. Computerized diagnosis tools (CAD) are used to create a classification between natural and abnormal lung tissue that may improve the ability of the radiologist \cite{JAVAID2016125,ABBAS2017325}.
The onset of lung cancer begins in the lungs, while secondary lung cancer begins elsewhere in the body and reaches the lungs. A pulmonary nodule is an oval or round growth in the lungs. The size of the nodules varies from a few millimeters to 5 centimeters. Given the shape and size of the nodule, classifying is a challenging task. Detection of large-sized malignant nodules is straightforward, but we have difficulty identifying small malignant nodules \cite{CHOI201437}.
The lung cancer death rate has weakened by 45\% since 1990 in men and by 19\% since 2002 in women due to cutbacks in smoking, with the pace of decline quickening over the past decade; from 2011 to 2015, the rate decreased by 3.8\% per year in men and by 2.3\% per year in women \cite{7375045}. According to Cancer Research UK, the five-year survival rate for patients diagnosed in stage one is more than 55\%, while the survival rate in patients with lung cancer in stage four is almost 5\% \cite{XIE2021100907}.
Computer-aided diagnosis systems (CAD) are effective schemes for identifying and detecting various pulmonary lesions. The main purpose of these systems is to assist the radiologist in various stages of diagnosis. The CAD system output acts as the second opinion for radiologists before the final diagnosis. In this way, researchers are developing more auto CAD systems for lung cancer. Many different publications have provided auto nodule detection systems using image processing, including various features extraction, classification, and segmentation techniques.

\section{Related works}
The diagnosis of pulmonary lesions is a very important topic, and a lot of work has been done in this field. However, due to many different types of pulmonary lesions and difficulty of diagnosis in this field, they are constantly seeking to increase the accuracy of existing systems. Thus, we have tried to design and train a more accurate system than the existing systems.
Various existing works are as follows:
Ying Xie et al. applied an interdisciplinary mechanism based on metabolomics and six machine learning methods and reached the sensitivity of  98.1\%, AUC in 0.989, and Specificity in 100.0\%. As a result, the machine learning methods are AdaBoost, K-nearest neighbor (KNN), Naive Bayes, Support Vector Machine (SVM), Random Forest, and Neural Network. They also recommended Naive Bayes as a suitable method \cite{MAGALHAESBARROSNETTO20121110}. Also, Netto et al. worked on the automatic separation of pulmonary nodules with growing neural gas and SVM. Their purpose was to automatically collect lung nodules through computed tomography images using the growth neural gas (GNG) algorithm to isolate structures with very similar properties to the lung nodules. They then used the distance conversion to separate the partitioned structures that connect the blood vessels and bronchitis. Finally, they used a set of features of the shape and texture using the SVM classifier to classify these structures: lung nodules \cite{4665129}.
Additionally, Lee et al. used a collection of classifiers called random forest in two stages. The first stage was diagnosing lung nodules and the second stage was false positive reduction \cite{yegraph}.
Moreover, Yi et al. worked with five attributes, including intensity information, shape index, 3D space, and location. They worked more on segmentation issues and reached 81\% accuracy \cite{simonyan2014very}.
Javid et al. worked on identifying nodes such as heart and muscle nodes. A brief analysis of CT histograms is performed to select an appropriate threshold for better results. A simple morphological closing is used in the segmentation of the lung area. The K-means clustering is applied for the initial detection and segmentation of potential nodes. This segmentation eventually reached a sensitivity of 91.65\% \cite{JAVAID2016125}.

\section{Material}
To train and evaluate the model presented in this article, we have been preparing a dataset, including CT-scan images of the local patients' pulmonary lesions. More than 10000 slides were used to prepare this dataset. All images were tagged and classified by specialist physicians. There are a large number of various types of lesions in this dataset.
Due to the large size of the model, there is a need for the right hardware resources to train these models, hardware resources, and sufficient time was provided, and all steps were successfully completed.

\section{Proposed Method}
Our proposed model is a complete and automatic system for the classification and segmentation of pulmonary lesions. In the first stage, a deep convolutional neural network automatically extracts features from CT-scan images. In the second stage, based on extracted features, an Ensemble Gradient boosting classifier identifies pulmonary lesions. Finally, in the third stage, CT-scan images are segmented by a hybrid fuzzy clustering – Level set method based on \cite{LI20111}.
Among the proposed system's positive features, the reduction of diagnosis time and high accuracy can be noted. The proposed method is a complete CAD system. After receiving the image, it automatically performs all the steps of extracting features, classification, and segmentation and provides the final output.
Extracting a feature involves extracting a higher level of information from raw pixel values that differentiate between different categories. Based on specific algorithms such as HOG, Haar, SIFT, LBP, GIST, the image features are extracted in classic methods. Based on these features extraction, the classifier is trained, and the final model is achieved. Some of these classification modules like SVM, logistic regression, random forest, KNN can be noted. One of the problems of classic methods is choosing and designing a suitable feature extraction method. A variety of different methods have been proposed over the years. Each method often works well in a particular field, and for new issues, there is a need to improve and change the feature extraction technique.
The problem with traditional methods is that the feature extraction method cannot be set based on classes and images. Therefore, if the selected feature does not have the abstract needed to identify the categories, regardless of the type of classification strategy used, the classification model's accuracy will be very low.  The problem with classic methods is always to find a distinctive feature among several features. Also, achieving accuracy, such as human accuracy, has been a big challenge. That is why it took years to have a flexible computer vision system (such as OCR, face recognition, image categorization, and object recognition) that works with various data. Another problem with these methods is that it is entirely different from how we learn to recognize things. Immediately after the birth of a child, he cannot understand his surroundings, but with the advancement and processing of data, he learns to identify things. This philosophy is behind deep learning. A computational model is trained in deep learning based on existing datasets and then automatically extracts the best features. Most deep neural networks require a lot of memory and computation, especially when training. Hence, this is an essential concern in these networks.

\subsection{VGG Convolutional Neural Network}
\begin{figure}
  \centering
  \includegraphics[scale=0.5]{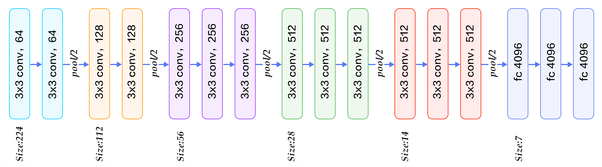}
  \caption{VGG Diagram \cite{simonyan2014very}}
  \label{VGGdiagram}
\end{figure}

At present, convolutional neural networks have managed to surpass humans in computer vision tasks, such as image classification. The image classification means determining which image belongs to which class. In the VGG network \cite{simonyan2014very}, for the first time, they used tiny $3 \times 3$ filters in each convolution layer and also combined them as a sequence of convolutions. Contrary to the principles of LeNet, which uses large convolutions to capture similar features in an image, as well as AlexNet, which uses $9 \times 9$ or $11 \times 11$ convolution filters, filters in the VGG network begin to shrink and approach The bad $1 \times 1$ convolution that LeNet wanted to avoid. These ideas are also used in newer architectures such as Inception and ResNet \cite{simonyan2014very,ZOU2019266}. Illustrated the VGG Diagram in Figure. \ref{VGGdiagram}.

\subsubsection{Specification and Parameters}
For the first convolutional layer, the network must learn $64$ filters of size $3 \times 3$ with input depth $(3)$. In addition, each of the $64$ filters has a bias, so the total number of parameters is $64 \times 3 \times 3 \times 3 + 64 = 1792$. It can be applied the same logic to other convolutional layers.
The depth of an output layer will be the number of convolution filters. The padding is selected as 1 pixel, so the spatial resolution is maintained through the convolutional layers. Thus, the spatial resolution will only change at the pooling layers. Therefore, the first convolutional layer's output will be $64 \times 224 \times 224$.
The pooling layer does not learn anything, so we have $0$ learning parameters. To calculate the pooling layer's output, we need to consider the size of the window and the step.
To calculate the number of parameters in fully connected layers, we must multiply the number of units in the previous layer with the current layer's number of units. By following the previous paragraph's logic, it can be seen that the number of units in the last convolutional layer will be $512 \times 7 \times 7$. Therefore, the total number of parameters in the first fully connected layer is $7 \times 7 \times 512 \times 4096 + 4096 = 1027645447$ \cite{simonyan2014very, 726791, ZOU2019266}.
As already mentioned, this network consists of two parts: a convolutional section and a fully connected section. The first $5$ blocks form the convolution section that is responsible for the feature extraction. The fully connected section consists of three dense layers that perform Classification.

\subsubsection{Transfer Learning}
\begin{figure}[h!]
  \centering
  \includegraphics[scale=0.6]{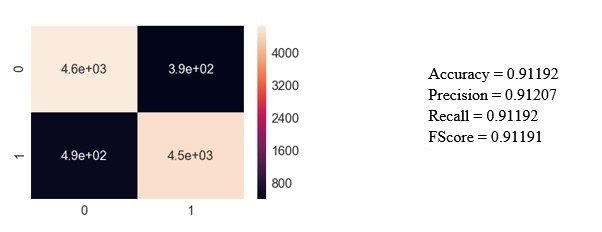}
  \caption{Transfer Learning Results}
  \label{transfer learnig results}
\end{figure}

In the dataset we provide, we encounter many intra-class patterns for positive samples, so we need to have a model that can learn these intra-class different patterns. The VGG network can understand complex models because of the high number of learnable parameters. Hence, if we have enough data and time, and proper hardware to train this network, we can achieve high accuracy.
First, the default fully connected part of the VGG network is removed, and the required fully connected layers are added. These layers are arranged after the convolution layers, respectively. Also, to increase the model's generalization and prevent overfitting, a dropout layer of $0.1$ is used. In practice, we use this layer to increase the accuracy of the model on the test data.
In the modified network, first, the convolution layers were frozen. Then we began to train the fully connected layers due to the possibility of a large difference in the initial values of fully connected layer weights from optimal values. If the convolutional layer is trained, it may be possible to reduce the system's absolute accuracy by inappropriately changing convolution weights. Hence, it is just the training of the fully connected layer at this stage, and we do not change the weights of the convolutional layers \cite{li2017hybrid}. The results are as follows: (Figure. \ref{transfer learnig results} shows the confusion matrix of these results).

\begin{figure}
  \centering
  \includegraphics[scale=0.5]{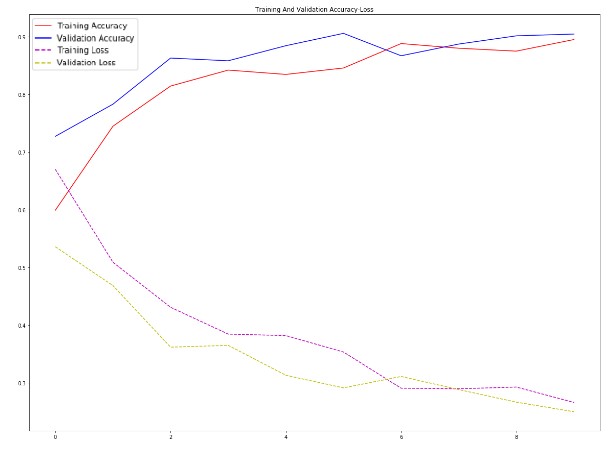}
  \caption{Transfer Learning / Training and Validation / Accuracy and Loss}
  \label{trnasfer learning training and validation accuracy and loss}
\end{figure}
As seen in Figure. \ref{trnasfer learning training and validation accuracy and loss}, almost the loss and accuracy are reached relative stability and no longer change after ten epochs. Continuing training in these situations may lead to network overfitting, so the training ends.

\subsubsection{Fine-tuning}
\begin{figure}
  \centering
  \includegraphics[scale=0.5]{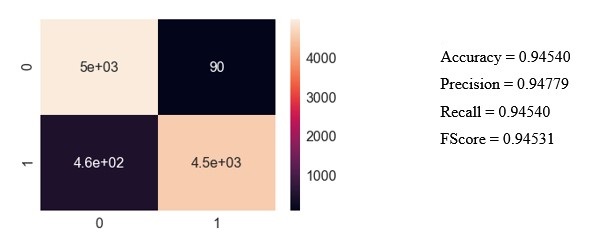}
  \caption{Fine-tuning Results}
  \label{fine tuning results}
\end{figure}

In the next step, we proceeded to Fine-tuning of the network so that only the weights of the last two blocks and the fully connected layer are trained \cite{tan2018survey}. Figure. \ref{fine tuning results} shows the results of the Fine-tuning step.
The last two blocks of the VGG network include convolutional modules and pooling modules. By training these modules, we see an increase in the accuracy of the results.

\begin{figure}
  \centering
  \includegraphics[scale=0.4]{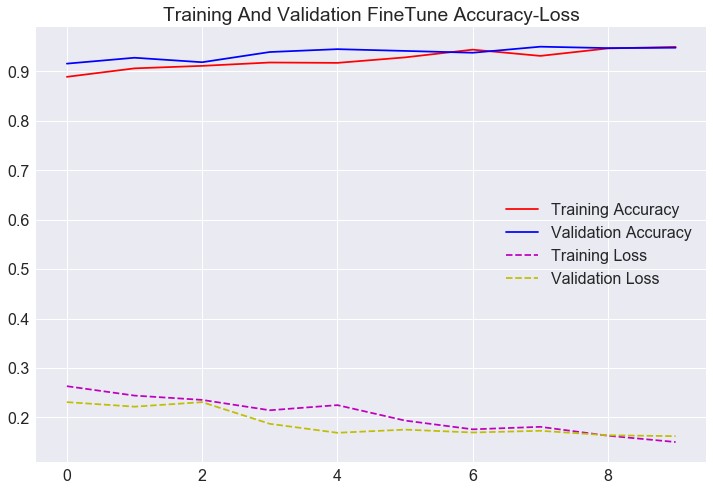}
  \caption{Fine-tuning / Training and Validation / Accuracy and Loss}
  \label{fine tuning training and validation accuracy and loss}
\end{figure}
As can be seen in Figure. \ref{fine tuning training and validation accuracy and loss}, after about ten epochs, loss and accuracy changes are significantly reduced and converged. Therefore, at this stage, we stopped training.

\subsubsection{Feature Extraction}
Our goal is to use the convolutional VGG network as an automatic and accurate feature extraction method. One of the ideas used in this article is to use an Ensemble-based Gradient boosting classifier instead of a fully connected layer to increase the accuracy of the pulmonary lesions' classification. Hence, we extracted the features from the CNN layers of the VGG model, and instead of the fully connected layer, we used classifiers based on Ensemble and Gradient boosting methods. What we will see below is the achievement of higher accuracy with this technique.

\subsection{Classification}
At this stage, different classifiers were trained based on the extracted features of the VGG convolutional neural network. Then the accuracy of each of these models was calculated on the test data. The results of these models are presented below.

\subsubsection{Ensemble Methods}

The ensemble methods are a concept of machine learning, the main idea of using multiple models to create a single and better algorithm. In other words, the methods in which multiple classifiers combine to make a more robust model. The accuracy of the generated model is greater than the accuracy of each of the initial models. One method for combining the results of classifiers is the majority vote. Voting and averaging are two of the easiest methods in the ensemble methods. Each of these methods has a simple understanding and implementation, voting is for classification, and averaging is for regression. In both ways, the first step is to create multiple classification/regression models using some of the training datasets. Each base model can be created using different training dataset divisions and the same algorithm or using the same dataset with different algorithms, or any other method \cite{demir17}.

\subsubsection{Boosting and Bagging}
Bagging and boosting are both algorithms of ensemble methods \cite{GONZALEZ2020205}, which combine a set of poor learners to create a strong learner who performs better. The leading cause of the error is related to noise, bias, and variance. Ensembles help reduce these factors. These methods are designed to improve the stability and accuracy of machine learning algorithms. The use and combination of several classifiers reduce the final model’s variance, especially for unstable classifiers, and may produce a more reliable model.
In Bagging, each element has a similar probability in a new dataset. Nevertheless, in boosting, the elements are weighed to increase the impact, and therefore some will be more involved in the training process. In Bagging, the training phase is parallel (for example, each model is built independently), but in Boosting, the new learner is sequentially arranged.
In boosting algorithms, each classifier is trained on the dataset based on the previous classifiers’ success. After each training step, weights are distributed. Data that is classified incorrectly increases its weight so that the classifier is forced to focus on that. To predict the class of new data, we only need to apply the learners to the newly observed data. Bagging results are obtained by averaging the responses of all learners (or the majority vote). However, in boosting, the second set of weights is allocated to learners to obtain a weighted average of all classifiers’ results.
At the boosting training stage, the algorithm assigns weights to each model. A classifier with good results gets a higher weight than a weak classifier. So boosting also needs to keep track of learners’ errors.
Boosting includes three simple steps:

\begin{itemize}
  \item A basic $F0$ model is defined for predicting the target variable $y$. This model is associated with a residual value $(y-F0)$.
  \item A new $H1$ model is fitted on the residual of the previous stage.
  \item Now, $F0$ and $H1$ are combined for $F1$ (the boosted version of $F0$). The average square error of $F1$ will be less than $F0$:
\end{itemize}

\begin{equation}
F_1(X)= -F_0(X) + H_1(X)
\end{equation}

To improve the performance of $F1$, we can create a new $F2$ after the residual of $F1$.

\begin{equation}
F_2(X)= -F_1(X) + H_2(X)
\end{equation}

This can be done for $'m'$ iterations until the residual value reaches our lowest target value.	
\begin{equation}
F_m(X) < -F_{m-1} (X) + H_m(X)
\end{equation}

As a first step, the model must begin with a function $F0(x)$. $F0(x)$ must be a function that minimizes the loss function or MSE\footnote{Mean Square Error}, in this case:

\begin{equation}
F_0(x) = argmin_Y\displaystyle\sum_{i=1}^{n} L(Y_i - y)
\end{equation}

\begin{equation}
argmin_Y \displaystyle\sum_{i=1}^{n} L(Y_i - y) = argmin_Y \displaystyle\sum_{i=1}^{n} (Y_i - y)^2
\end{equation}

\subsubsection{XGBoost}

XGBoost is similar to Gradient boosting algorithm, but it has a few tricks up its sleeve, making it stand out from the rest. It has proven itself in terms of performance and speed. XGBoost unlike other GBM\footnote{Glioblastoma} methods that first specify the step, and then the step value, directly determine the step using the following statement for each x in the data:

\begin{equation}
\frac{\delta L(y‚f^((m-1))(x) + f_m (x))}{\delta f_m (x)} = 0
\end{equation}

By doing second-order Taylor expansion of the loss function around the current estimate $f_{m-1}(x)$, we get:

\begin{multline}
L \left (y‚ f^{m-1}(x) + f_m(x)\right) \\
\approx L(y‚f^{m-1}(x)) + g_m(x) f_m(x) + \frac{1}{2}h_m(x)f_m(x)^2
\end{multline}

where $g_m(x)$ is the Gradient, same as the one in GBM, and $h_m(x)$ is the Hessian (second order derivative) at the current estimate.

\begin{equation}
h_m(x) = \frac{\delta^2 L(Y, f(x))}{\delta f(x)^2}
\end{equation}

\begin{equation}
f(x) = f^{m-1}(x)
\end{equation}

Then the loss function can be rewritten as:

\begin{multline}
L(f_m) \approx \displaystyle\sum_{i=1}^{n}[g_m(x_i)f_m(x_i) + \frac{1}{2} h_m(x_i)f_m(x_i)^2] + const. \\
\propto \sum_{j=1}^{T_m} \sum_{i\in R_jm}^{0}\left[g_m(x_i)w_jm + \frac{1}{2}h_m(x_i)w^2_jm\right]
\end{multline}

While Gradient Boosting follows negative Gradients to optimize the loss function, XGBoost uses Taylor expansion to calculate the value of the loss function for different base learners. XGBoost does not explore all possible tree structures but builds a tree greedily, and its regularization term penalizes building a complex tree with several leaf nodes \cite{chen2016xgboost,saucecat}.

\subsubsection{Classification Results}

\begin{figure}
  \centering
  \includegraphics[scale=0.6]{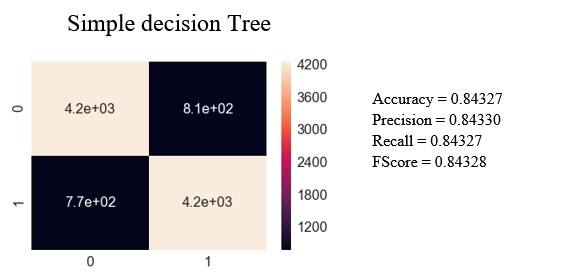}
  \caption{Simple Decision Tree Confusion Matrix}
  \label{simple decision tree confusion matrix}
\end{figure}

\begin{figure}
  \centering
  \includegraphics[scale=0.6]{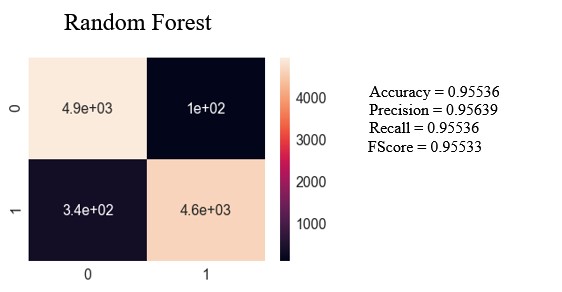}
  \caption{Random Forest Confusion Matrix}
  \label{Random Forest Confusion Matrix}
\end{figure}

\begin{figure}
  \centering
  \includegraphics[scale=0.6]{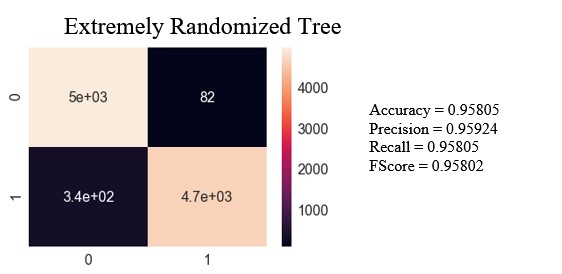}
  \caption{Extremely Randomized Tree Confusion Matrix}
  \label{Extemely Randomized Tree Confusion Matrix}
\end{figure}

\begin{figure}
  \centering
  \includegraphics[scale=0.6]{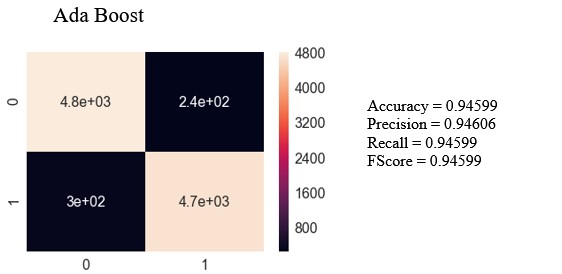}
  \caption{AdaBoost Confusion Matrix}
  \label{AdaBoost Confusion Matrix}
\end{figure}

\begin{figure}
  \centering
  \includegraphics[scale=0.6]{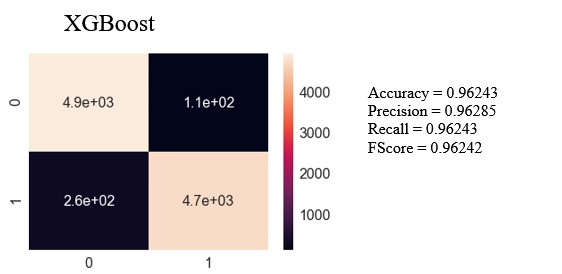}
  \caption{XGBoost Confusion Matrix}
  \label{XGBoost Confusion Matrix}
\end{figure}

\begin{figure}
  \centering
  \includegraphics[scale=0.6]{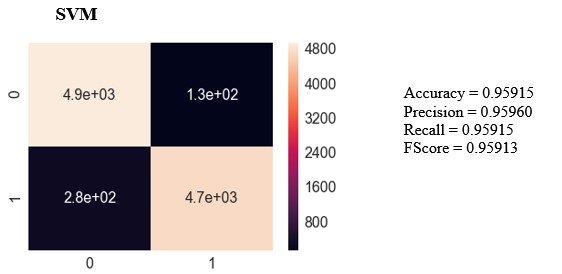}
  \caption{SVM Confusion Matrix}
  \label{SVM Confusion Matrix}
\end{figure}

In the first try, we used the output of the last layer of the CNN as the feature vector, which contains 25088 features. In the second try, instead of using the output of the last layer of CNN, we used the output of the first Dense layer which achieved a higher final accuracy. The results are as follows: (Figures. \ref{simple decision tree confusion matrix} to \ref{SVM Confusion Matrix})

As can be seen, the XGBoost classifier's accuracy is 96\%, and the accuracy of the fully-connected layer is 94\%, so we showed that using the XGBoost classifier instead of the dense layers of the VGG network leads to higher accuracy \cite{li2017hybrid, ren2017novel}.

\subsection{Pulmonary Lesions Segmentation}

Segmentation is the process of partitioning an image into different meaningful segments. These segments often correspond to different tissue classes, organs, pathologies, or other biologically relevant structures in medical imaging. Medical image segmentation is made difficult by low contrast, noise, and other imaging ambiguities \cite{7375045}. A major difficulty of medical image segmentation is the high variability in medical images. The result of the segmentation can then be used to obtain further diagnostic insights.
Level set methods, based on partial differential equations (PDEs), are effective in the medical image segmentation tasks. However, to use this method, determining its control parameters is very important. Hence, other methods are used to determine these parameters. One of these techniques is FCM clustering. Using this method, with the medical image's initial segmentation into several clusters, the level set initial parameters can be set automatically \cite{4618616,FOROUZANFAR2010160,SWIERCZYNSKI201858,4535886}.

\subsubsection{Fuzzy C-Means Clustering}

The fuzzy C-means algorithm is very similar to the k-means algorithm. The steps are as follows:

\begin{itemize}
    \item Choose a number of clusters.
    \item Assign coefficients randomly to each data point for being in the clusters.
    \item Repeat until the algorithm has converged
        \begin{itemize}[label $=0$]
            \item Compute the centroid for each cluster.
            \item For each data point, compute its coefficients of being in the clusters.
        \end{itemize}
\end{itemize}

Each point $x$ contains a set of coefficients that determine the membership degree in the k-th cluster, $w_k(x)$. In Fuzzy C-means, the center of each cluster is averaged from all points. Weighing the membership degree is obtained by using the following formula:

\begin{equation}
C_k = \frac{\sum_{x}^{W_k(x)^m x}}{\sum_{x}^{W_k(x)^m}}
\end{equation}

$m$ is a hyperparameter that controls how clustering works.
The goal of the Fuzzy C-means algorithm is to minimize the following target function \cite{ZHANG200437}:

\begin{equation}
C_k = \frac{\sum_{x}^{W_k(x)^m x}}{\sum_{x}^{W_k(x)^m}}
\end{equation}

\begin{equation}
argmin_c \displaystyle\sum_{i=1}^{n}\displaystyle\sum_{j=1}^{c}W_{ij}^m \displaystyle\left\lvert\lvert x_i - c_j \right\rvert\rvert^2
\end{equation}

where:
\begin{equation}
W_{ij} = \frac{1}{\displaystyle\sum_{k=1}^{c}\frac{\left\lvert\lvert x_i - c_j \right\rvert\rvert^\frac{2}{m-1}} {\left\lvert\lvert x_i - c_k \right\rvert\rvert}}
\end{equation}

\subsubsection{Level Set Methods}

The idea of extending a surface ($\Phi$) instead of a front boundary (C) is used in this method, and the front boundary is defined so that all points with no elevation ($\Phi = 0$). When the surface evolves and develops, the surface with a zero level set takes on various shapes. The surface $\Phi$ points and our reference surface form our implicit boundary, and the zero level set shows contours splitting and merging. In this method, additional care is not required for topological changes. Therefore, this method is more suitable for our application.

\subsubsection{The mathematical study of Level Set Methods}

Assume that the point $x =(x, y)$ belongs to the evolving front. So it changes over time, and $x(t)$  is the position over time. At any time $t$, for each point $x(t)$ on the front, the surface has by definition no height, thus:

\begin{equation}
\phi (x(t), t) = 0
\end{equation}

To obtain the boundary, we require zero on Level Set, within the fact that it could carry any value. Assuming a primary $\phi$ at $t = 0$, we may obtain $\phi$ at any time $t$ with the equation of motion $\delta\phi / \delta t$. Based on the following chain rules, we have:

\begin{equation}
\frac{\delta\phi(x(t), t)}{\delta t} = 0
\end{equation}

\begin{equation}
\frac{\delta\phi}{\delta x(t)}\frac{\delta x(t)}{\delta t} + \frac{\delta t}{t}\frac{t}{t} = 0
\end{equation}

\begin{equation}
\frac{\delta\phi}{\delta x_t}x_t + \phi_t = 0
\end{equation}

We call here $\delta\phi/\delta t = \nabla\phi$. Also, $x(t)$ is obtained through the force $F$, which is normalized to the surface, and therefore:

\begin{equation}
x_t = F\left(x(t)\right)n
\end{equation}

Such that  $n = \nabla\phi/\lvert\nabla\Phi\rvert$ and the previous moving equations are rewritten as follows:

\begin{equation}
\phi_t + \nabla\phi x_t = 0
\end{equation}

\begin{equation}
\phi_t + \nabla\phi F_n = 0
\end{equation}

\begin{equation}
\phi_t + F\nabla_\phi\frac{\nabla_\phi}{\lvert\nabla_\phi\rvert} = 0
\end{equation}

\begin{equation}
\phi_t + F\lvert\nabla_\phi\rvert = 0
\end{equation}

The last equation is the $\phi$ moving equation. If $\phi$ is given at time $t = 0$ and its motion equation is known over time, it is now possible to find $\phi(x, y, t)$ any time $t$ through the expansion of $\phi$ over time. An interesting feature with $\phi$ is that we can find the curvature of the curve by the following equation:


\begin{equation}\label{equ24}
k = \nabla\frac{\nabla_\phi}{\lvert\nabla\phi\rvert}= \frac{\phi_{xx}\phi_y^2 - 2\phi_{xy}\phi_x\phi_y + \phi_{yy}\phi_x^2}{(\phi_x^2+\phi_y)^\frac{1}{2}}
\end{equation}

This parameter can be used to control the smoothness and uniformity of the evolving front boundary \cite{Lombaert26,OSHER198812,kass1988snakes}.

\subsubsection{Integrating spatial fuzzy clustering with Level Set}

In this section, we used a method based on the work of Li et al. for pulmonary lesions image segmentation \cite{LI20111}.
The level set method requires initial control parameters and sometimes also requires manual intervention to control these parameters. In this paper, a Fuzzy level set method is used to facilitate the segmentation of pulmonary lesions. Level set evolution can be started directly from the primary segmentation by spatial fuzzy clustering. Control of the parameters of the level set also evolves according to the fuzzy clustering results. Such methods help to better segmentation.

This method uses fuzzy clustering as the initial surface function. The FCM algorithm with spatial information can accurately estimate the boundaries. Therefore, the level set evolution will begin from an area close to the actual boundaries. Here, information from fuzzy clustering is used to estimate control parameters, which reduces manual interventions. The new level set fuzzy algorithm automates the initial settings and parameter setup of the level set using local fuzzy clustering. This is an FCM with spatial constraints to determine the approximate lines of interest in a medical image \cite{4618616}.

The model evolution equation presented by Li et al. is as follows:

\begin{equation}\label{equ25}
\epsilon(g,\phi) = \lambda\delta(\phi)div(g\frac{\nabla\phi}{\lvert\nabla\phi\rvert}) + gG(R_k)\delta(\phi)
\end{equation}

where, $\lambda$ is the Coefficient of the contour length for smoothness regulation,
$\epsilon$ is Regulator for Dirac function $\delta(\phi)$,
$\phi$ converts the 2D image segmentation problem into a 3D problem,
$(\delta(\phi))$ denotes the Dirac function, $\epsilon(g,\phi)$ (attracts $\phi$ towards the variational boundary, which is similar to the standard level set methods.

\begin{equation}\label{equ26}
R_k = \lbrace r_k = \mu_{nk}, n = x \times N_y + y \rbrace
\end{equation}
$R_k$ is the component of interest in FCM.

Finally, surface evolution can be regulated using local fuzzy clustering. In other words, the level set evolution is stable when it approaches the actual boundaries, which not only prevents boundary leakage but also prevents manual intervention. All these improvements lead to a strong algorithm for medical image segmentation.

\subsection{Proposed system's final outputs and results}

\begin{figure}[h!]
  \centering
  \includegraphics[scale=0.7]{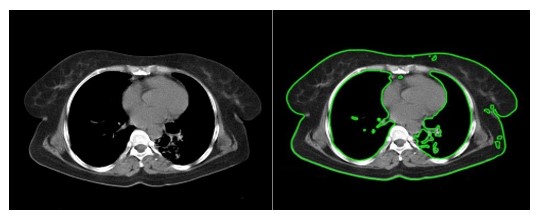}
  \caption{Left picture: Positive Sample - Right: Segmented Image}
  \label{13 Left Positive Sample Right Segmented Image}
\end{figure}

\begin{figure}[h!]
  \centering
  \includegraphics[scale=0.7]{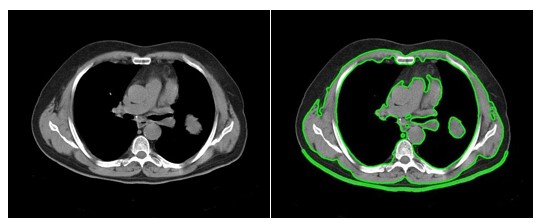}
  \caption{Left: Positive Sample - Right: Segmented Image}
  \label{14 Left Positive Sample Right Segmented Image}
\end{figure}

After reading the CT-Scan image, its features are automatically extracted by the VGG convolutional neural network. Then, based on extracted features, it is classified using the XGBoost classifier. In the next step, if there is a lesion in the CT-scan image, the observed lesion is segmented by the mentioned hybrid segmentation method. A number of system outputs are shown below in Figures. \ref{Left picture is the Positive Sample and Right is Segmented Image} to \ref{14 Left Positive Sample Right Segmented Image}.

\begin{figure}[h!]
  \centering
  \includegraphics[scale=0.7]{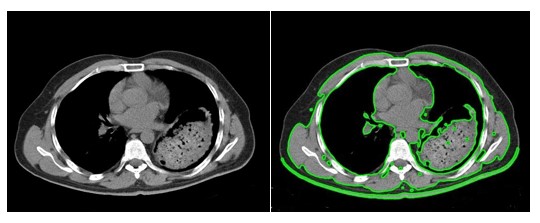}
  \caption{Left picture is the Positive Sample and Right is Segmented Image}
  \label{Left picture is the Positive Sample and Right is Segmented Image}
\end{figure}

\section{Discussion and Conclusion}

The model proposed in this article is based on the latest methods of artificial intelligence and deep learning. To increase the accuracy, we used several hybrid methods for required tasks and the result was very successful.
In the proposed method, first, a convolutional neural network was used to automatically extract the best features. Duo to a large number of different intra-class patterns, the VGG network was chosen for this work. The VGG network has many learnable parameters so it can learn many different patterns very well. With these extracted features various classifiers were trained and tested. The best results were achieved with the XGBoost classifier. And we showed that the use of ensemble and Gradient boosting based classifiers instead of a fully connected layer in the VGG network, increases the accuracy of the classification of pulmonary lesions. Then, in the case of a positive diagnosis of a pulmonary lesion, a hybrid fuzzy level set method (Li et al) was used for image segmentation. A dataset including CT-scan images of patients in Mashhad Local Area was collected and labeled by a specialist. We used this dataset for training and testing the proposed models.
Significant features of the proposed system include the followings:
\begin{itemize}
  \item Using a deep convolutional model to automatically extract the best features.
  \item The use of the VGG network based on fully-connected a large number of different intra-class patterns (Due to a large number of different types of pulmonary lesions.
  \item Applying XGBoost instead of a fully connected layer.
  \item Using a highly accurate and hybrid method for segmentation of pulmonary lesions.
  \item Designing an automated and complete system as a health care CAD system.
  \item Achieving very high accuracy so that the proposed model can be operationally used in the health care industry.
  \item Using a local dataset based on native patients in Mashhad.
\end{itemize}

Future recommendations for this work include: Determine the exact type of lesion, determine the risk of a diagnosed lesion and extend the dataset so that it covers all pulmonary lesions. Providing such dataset requires a great deal of time and money. To achieve this, there is a need for financial and scientific support in the form of an interdisciplinary research team. Diagnosis of pulmonary lesions due to the very high diversity is a hard and highly specialized task. Therefore, the preparation of such systems requires high technical knowledge. With the advancement of artificial intelligence methods, the accuracy of such systems can always be improved. The field of advancing and improving these systems will always be open to researchers.
\section*{Acknowledgment}
We would like to thank Dr. Mahjoub and \href{http://behsazteb.ir/en/page/show/Home.html}{Behsazteb Medical Imaging Center} for providing local CT-scan images of patients with human lung problems and providing diverse comments and reports on these images, which enabled us not only to categorize them accurately but also undoubtedly had a significant impact on this work's overall improvement.

\ifCLASSOPTIONcaptionsoff
  \newpage
\fi

\bibliographystyle{ieeetr}
\bibliography{REFRENCES}


\appendices
\section{Graphical Abstract}
\begin{figure}[h!]
  \centering
  \includegraphics[scale=0.4]{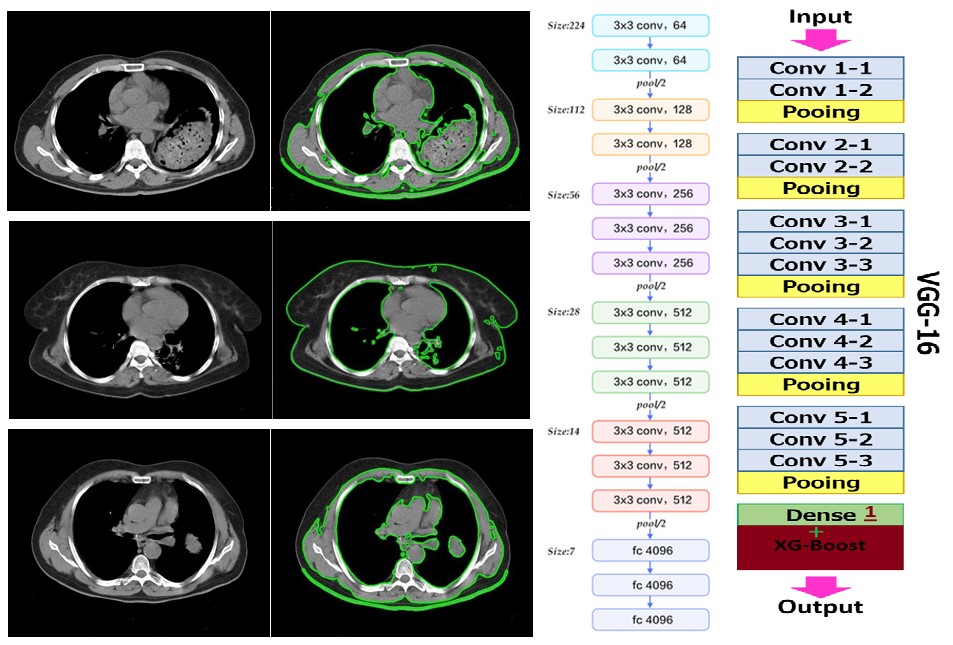}
  \caption{Graphical Abstract}
  \label{GraphicalAbstract}
\end{figure}

\section{Highlights:}
\begin{itemize}
  \item A pulmonary lesion classification method with slightly more than 96\% accuracy.
  \item Combining VGG and XG-Boost for lesion classification.
  \item Models are trained on a dataset with more than thousands samples.
  \item Gathering a new dataset of lung lesions based on local patients.
  \item Pulmonary lesion segmentation using an integrated Fuzzy Clustering-Level Set method.
  \item An automated lung lesion classification and segmentation method.
\end{itemize}

\end{document}